\documentclass[letterpaper, 10 pt, conference]{ieeeconf}  

\IEEEoverridecommandlockouts                              

\usepackage{graphics} 
\usepackage{epsfig} 
\usepackage{mathptmx} 
\usepackage{times} 
\usepackage{amsmath} 
\usepackage{amssymb}  

\usepackage{subfigure}
\usepackage{color}
\usepackage{booktabs}
\usepackage{pdfrender}
\usepackage{cite}

\usepackage{algorithmic}
\usepackage{array}
\usepackage{url}

\title{\LARGE \bf
Natural grasp intention recognition based on gaze fixation in human-robot interaction
}

\author{Bo Yang,~\IEEEmembership{Student~Member,~IEEE,}
Jian Huang$^{*}$,~\IEEEmembership{Senior~Member,~IEEE,}
Xiaolong Li,\\
Xinxing Chen,~\IEEEmembership{Student~Member,~IEEE,}
Caihua Xiong,~\IEEEmembership{Member,~IEEE} and
Yasuhisa~Hasegawa~\IEEEmembership{Member,~IEEE}
\thanks{This work was supported by the National Natural Science Foundation of China under Grant (No. U1913207) and the International Science and Technology Cooperation Program of China under Grant (No. 2017YFE0128300). \emph{(Corresponding author: Jian Huang)}}
\thanks{Bo Yang, Jian Huang, Xiaolong Li and Xinxing Chen are with the Key Laboratory of Image Processing and Intelligent Control, School of Artificial Intelligence and Automation, Huazhong University of Science and Technology, Wuhan 430074, China (e-mail:aleksibob@hust.edu.cn; huang\_jan@mail.hust.edu.cn; lixiaolong@hust.edu.cn; cxx@hust.edu.cn)}%
\thanks{Caihua Xiong is with the School of Mechanical Science and Engineering and the State Key Laboratory of Digital Manufacturing Equipment and Technology, Huazhong University of Science and Technology, Wuhan 430074, China (e-mail: chxiong@hust.edu.cn)}
\thanks{Yasuhisa Hasegawa is with Department of Micro-Nano Mechanical Secience and Engineering, Nagoya University, Furo-cho, Chikusa-ku, Nagoya 464‑8603, Japan (email: hasegawa@mein.nagoya-u.ac.jp)}%
}

\begin{document}

\maketitle
\thispagestyle{empty}
\pagestyle{empty}

\begin{abstract}

Eye movement is closely related to limb actions, so it can be used to infer movement intentions. More importantly, in some cases, eye movement is the only way for paralyzed and impaired patients with severe movement disorders to communicate and interact with the environment. Despite this, eye-tracking technology still has very limited application scenarios as an intention recognition method. The goal of this paper is to achieve a natural fixation-based grasping intention recognition method, with which a user with hand movement disorders can intuitively express what tasks he/she wants to do by directly looking at the object of interest.
Toward this goal, we design experiments to study the relationships of fixations in different tasks. We propose some quantitative features from these relationships and analyze them statistically. Then we design a natural method for grasping intention recognition.
The experimental results prove that the accuracy of the proposed method for the grasping intention recognition exceeds 89\% on the training objects. When this method is extendedly applied to objects not included in the training set, the average accuracy exceeds 85\%. The grasping experiment in the actual environment verifies the effectiveness of the proposed method. 

\end{abstract}


\section{Introduction}
Grasping is one of the important actions in people's daily life, but the limbs of some disabled people cannot implement the grasping function because of paralysis, amputation, etc. Wearable robots including exoskeletons~\cite{dollar2008lower}, prosthetics~\cite{belter2013mechanical}, and supernumerary robotic fingers~\cite{wu2015hold, hussain2016soft, ciullo2018analytical} can serve as alternative actuators.

Recognition of grasping intention is the premise for controlling wearable robots. The current main methods include EMG-based methods~\cite{salvietti2016compensating, marco2017surface}, EEG-based methods~\cite{penaloza2018bmi, downey2018intracortical}, and action-based methods~\cite{huang2015control, de2015recognizing}. Among these methods, the EMG-based method is not suitable for patients with hemiplegic paralysis, because the EMG signals of affected limbs have been distorted, from which it is difficult to identify the intention of patients. The accuracy and discriminability of EEG signals obtained by the noninvasive brain-machine interface are relatively low, and it is easy to be interfered by external factors. Although the invasive interface can obtain accurate EEG signals, most patients cannot accept craniotomy. The action-based methods use sensors such as airbags and IMUs to obtain the action states of the user's limb to infer the user's intention, but these methods are impossible for those patients who have lost their locomotivity.

Eye-tracking technology can be the feasible means of grasping intention recognition for such patients. The use of eye movement as a control interface relies on the fact that our gaze is proactive and directly correlated to action intentions and cognitions~\cite{hayhoe2005eye}. Furthermore, the oculomotor system control function is usually retained even in the most severe cases of paralysis related to muscular dystrophies or brain stroke. And in neurodegenerative diseases that affect the motor system (such as stroke, Parkinson's disease, or Multiple Sclerosis), it is easier for patients to control their gaze point rather than conducting stable skeletal movements~\cite{cipresso2011combined}.

Most of the studies using eye movement technology only focus on reaching tasks, e.g., Roni et al. used gaze controlling robotic arms to assist human hands in reaching tasks~\cite{maimon2017towards} , and Shaft et al. used eye movement controlling robotic arms to perform tasks\cite{shafti2019gaze}. Other studies have adopted unnatural methods to recognize the grasping intention of patients, such as asking users to deliberately blink or prolong the gaze, which will increase the cognitive load on users. It is not easy for patients with nerve damage to complete these unnatural actions.

In these case, natural human-robot interaction can be used to recognize the patients' intention. Natural human-robot interaction uses natural action to obtain the user's intention~\cite{ferland2013natural, valli2008design}, without imposing extra cognitive load on the users. This is similar to the user's control of their own limbs for activities without long-term training.

Previous studies have shown that there are differences in natural gaze between grasping and viewing behaviours~\cite{brouwer2009differences}. However, these studies focus on the effects of different objects on the first two fixations. To the best of our knowledge, there are only qualitative summaries of differences from experimental observations but no quantitative features that can be used for grasping intention recognition reported in existing literatures.

On this basis, we study the relationships of the fixations in the grasping tasks and viewing tasks when the thumb and index finger are visible, which is a common situation while humans are conducting grasping tasks. We analyze these relationships and extract some quantitative features to identify gasping intention.
Using the obtained features, we recognize the user's grasping intention and verify the effectiveness of the method. Our main contributions are as follows:

\begin{enumerate}[]
  \item We discover the relationships between the fixation selves, the fixations and objects, the fixations and grasping points in the grasping tasks and the viewing tasks. Compared with the viewing tasks, the fixations of the user are more centralized and tend to be close to the grasp point of the index finger in the grasping task.
  \item We propose four features from the relationships for grasping intention recognition. Then  we compare the accuracy of different features in recognition of grasping intention. On this basis, we implement a natural method to recognize the grasping intention by eye movement.
   \item We carry out grasping experiments with humanoid robotic arms in actual daily scenes, which verifies the effectiveness of the proposed method.
\end{enumerate}

\section{Methods}
\subsection{System overview and architecture}
\begin{figure*}[htbp]
  \centering
  \includegraphics[width=0.9\textwidth]{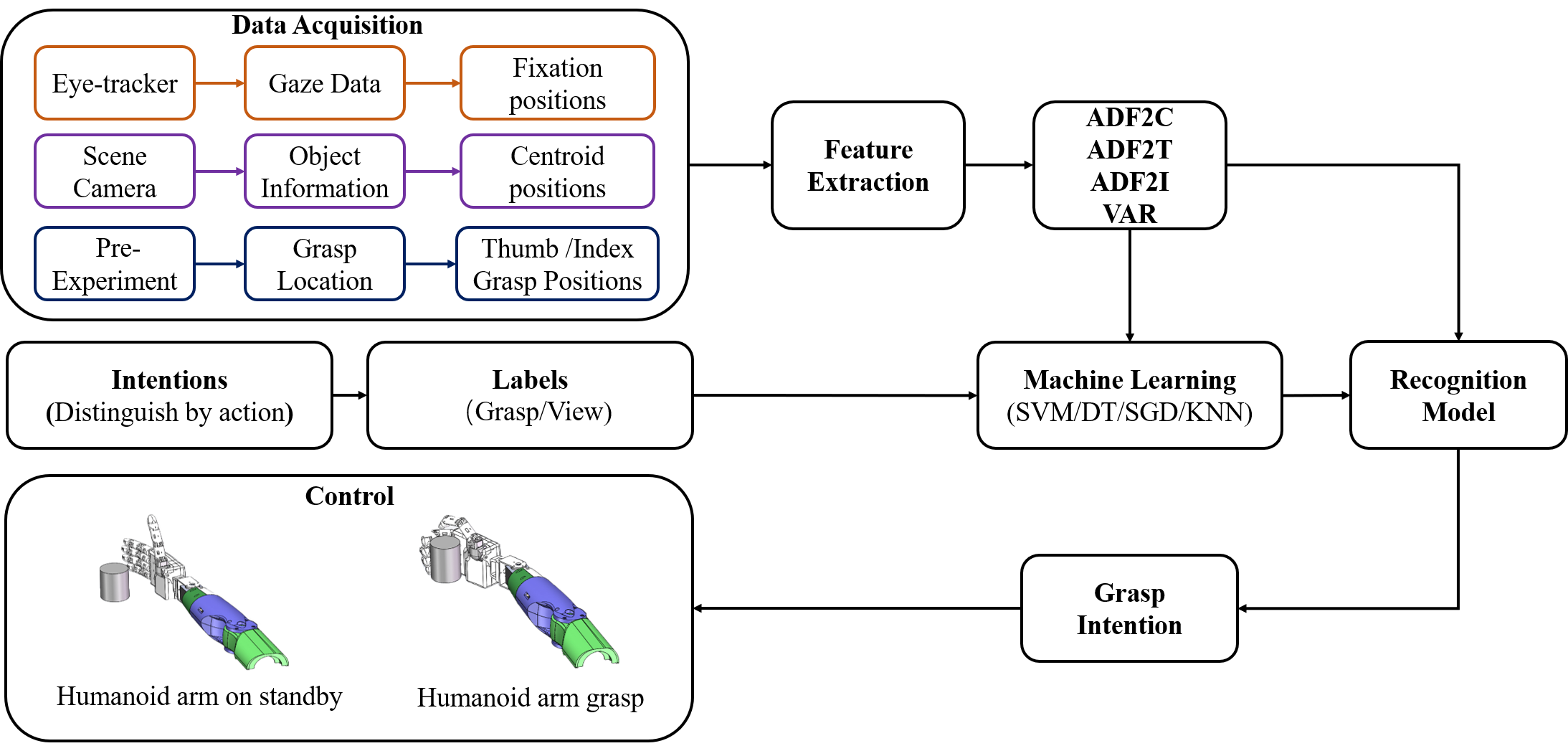}
  \caption{The block diagram of our system. The user's fixation positions are obtained by the eye-tracker, the object information is obtained by the image processing algorithms, and the grasping position is obtained by pre-experiment. The features proposed in this paper are extracted from these data and used to train the intent recognition model. In the task of intent recognition, the humanoid robot arm is controlled to complete the grasping action according to the model recognition result.}
  \label{block}
\end{figure*}

We aim to create a system to recognize grasping intention based on human natural gaze data. The system consists of  a binocular eye-tracker, a scene camera, a head stand and a black background (see Fig.~\ref{setup}). The block diagram in Fig.~\ref{block} depicts an overview of our system. Individual modalities are described in detail in the following.

\subsubsection{Binocular eye-tracker}
For 2D eye tracking, we use the Pupil Core~\cite{kassner2014pupil}.
The eye-tracker is built from two infrared (IR) cameras used to photograph the eyes, yielding a resolution of 192 $\times$ 192 pixels at a maximum frame-rate of 120 Hz. It has two extendable mounts for the infrared cameras, which can be adjusted to focus the IR cameras on each eye respectively for individuals. There are two IR LEDs attached to the cameras to illuminate the pupil of the user. Using the obtained infrared images, we can select the appropriate eye model and apply ellipse fitting parameters during system operation to detect the pupil of the user, and then use this information to estimate the user's fixation point.

An Intel Realsense D435 RGB-D camera (Intel Corporation, Santa Clara, California, USA) is mounted on the top of the eye-tracker as a scene camera, with a 3D printed frame. This camera is used to capture the user's ego-centric scene.

\subsubsection{Object centroid detection}
In the experiment environment, we use the OpenCV image processing library~\cite{kaehler2016learning} to process live scene video streams and detect the object's centroid in each frame with a contour-based algorithm.
To avoid the interference from ambient light, we use a solid color background. Besides, we filter the centroid of some disturbing objects based on the area of the object contour. Because it is impossible to detect the real centroid of 3D objects in RGB images, we use the 2D centroid of objects in images instead.

In real environment applications, we use Mask R-CNN~\cite{he2017mask} to quickly and accurately obtain object information. This machine learning-based method can effectively improve the robot system's ability to perceive the environment and identify objects that patients need to grasp in their daily lives.

\subsubsection{Grasp position detection}
In grasping tasks the thumb and the other fingers play different roles. The thumb guides the hand straight to the object's location, and then the other fingers represented by the index finger grip around the object to ensure a safe grasp~\cite{haggard1997hand}. We record the positions of the participants' digits to obtain the relationship between the fixation position and the grasping position. The grasping positions are obtained through pre-experiment.
In the preliminary experiment, we ask participants to grasp each object in repeated fashion and use the average positions as the grasping positions of the thumb and index finger.

\subsection{Experiment}
There are two categories of gaze behaviour: fixations and saccades. Fixations are those times when our eyes essentially stop scanning about the scene, holding the central foveal vision in place so that the visual system can take in detailed information about what is being looked at~\cite{rayner200935th}.
The main goal of the experiment is to investigate the difference of fixation modes of the participants under different task conditions of grasping and viewing.

Eight different participants carry out both viewing and grasping tasks. Their ages range between 20 and 30. Prior to the experiments, a short introduction is provided to these subjects, including technologies involved, the system setup and the research purpose. During the experiment, a head stand is used to fix the subjects' head. In practical applications, a head tracking module can be added to track the movement of the use's head, which is not covered in this paper.

\subsubsection{Procedure}
\begin{figure*}[thbp]
\centering
\subfigure[]{
\begin{minipage}[t]{0.45\linewidth}
\centering
\includegraphics[width=1\linewidth,height=3.8cm]{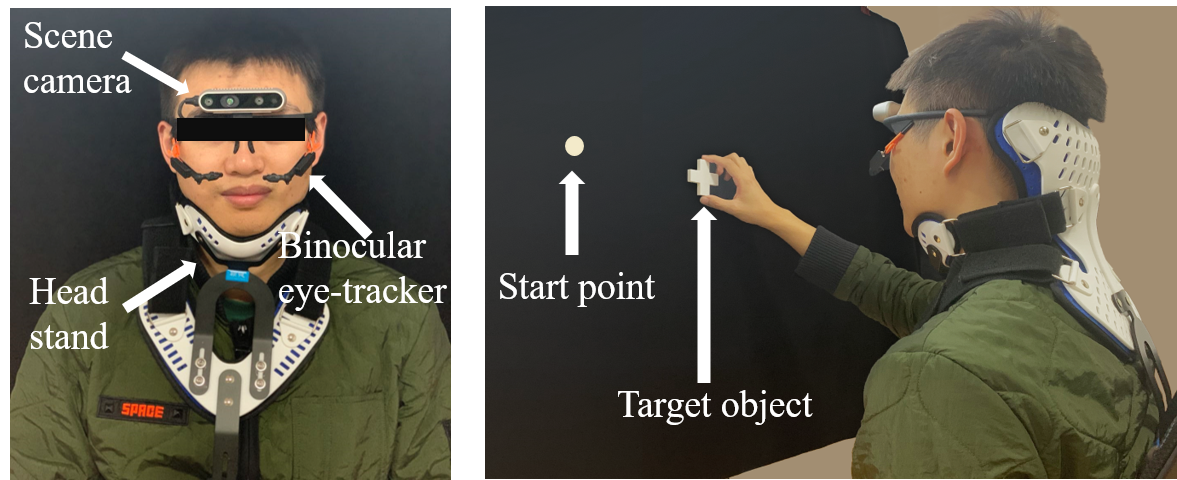}
\end{minipage}%
}%
\subfigure[]{
\begin{minipage}[t]{0.45\linewidth}
\centering
\includegraphics[width=1\linewidth]{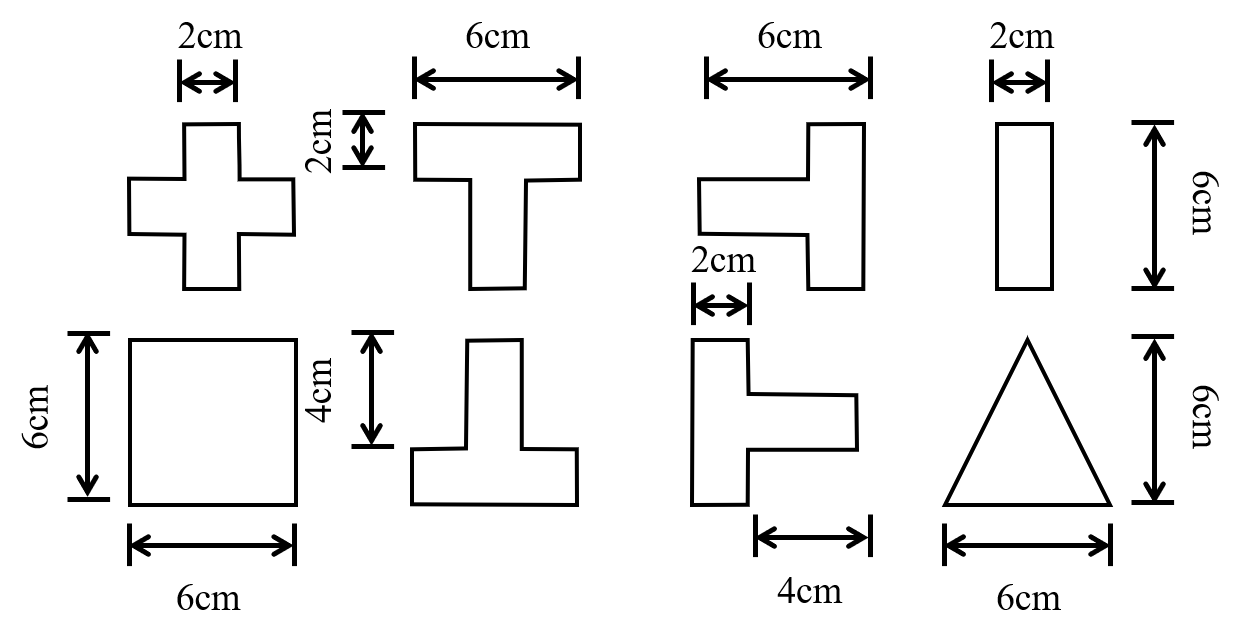}
\end{minipage}%
}%
\centering
\caption{The experimental scene and target shapes. (a) Schematic depiction of the setup and the way of grasping in the experiment. (b) Different shapes in the experiment.}
\label{setup}
\end{figure*}

The shapes that the participants are asked to grasp or view are magnetically attached to a black background, which is placed in front of an iron plate (Fig.~\ref{setup}). We use three different white plastic objects for training, depicted in Fig.~\ref{setup}(b): a square, a cross, a T-shape (presented in four different orientations).
They are 10 mm thick.
There are two types of grasping tasks, horizontal grasping and vertical grasping. The vertical T-shape is used for vertical grasping tasks, while the horizontal T-shape is used for horizontal grasping tasks. The square and cross shapes are used in both grasping tasks.
For each type of the grasping task, participants perform 4 shapes * 5 repetitions = 20 practice trials, presented in random order.
For the viewing task, participants perform the same number of trials.
We use two other shapes: a triangle and a bar (presented in 2 different orientations) for testing.

The eye-tracker needs to be calibrated before the experiment starts. Participants rest their head on a head stand, 50 cm in front of the monitor, which is the same distance as in the experiment. Both eyes are calibrated using a nine point calibration/validation procedure on the computer monitor. After the participants have performed half trials, the calibration is repeated. At the start of each trial, the participants fix at the start point and closes their eyes after fixation.

For grasping tasks, participants start a trial with the eyes closed and the hands resting on the lap. When hearing a ``grasp'' command, they open their eyes and immediately fix their eyes on the start point. Participants then stretch out their thumb and index finger to grasp the shape (Fig.~\ref{setup}(a)). Different grasping tasks have different grasping modes. The user's fingers of the horizontal grasping tasks contact with the left and right edges of the object, while the fingers of the vertical grasping tasks contact with the top and bottom edges.
The trial is ended when the ``finish" command is heard. After completing the trial, the participants return to the start point with their hands on the lap and their eyes closed, until hearing the verbal command to start next trial.

The procedure of the viewing task is similar to the grasping task, with the exception that instead of being asked to grasp, participants are asked to ``view the shape'' until hearing a "finish" command. The viewing process lasts for three seconds. The duration is measured by the participants through the pre-experiment. In the experiment, participants perform 10 grasping tests repeatedly and we select the maximum time of task execution as the benchmark time in the experiment.

\section{Data analysis}
Among 640 grasping and viewing trails, the data of 16 grasping trails (5\%) and 24 viewing trails (7.5\%) are rejected due to insufficient quality of eye data. We use the fixation detection algorithm to distinguish fixations from fast saccades. The parameters of the algorithm recommended by the manufacturer are set as follows:
\begin{equation}
\begin{aligned}
&dp{s_{\max }}{\rm{ = 3}}{\rm{.01}},~du{r_{\min }}{\rm{ = 80}}ms,~du{r_{\max }} = 400ms
\end{aligned}
\end{equation}
$dps_{max}$ indicates the threshold of the distance between all gaze locations during a fixation. $dur_{min}$ indicates the minimum duration in which the distance between every two gaze locations must not exceed $dps_{max}$ and $dur_{max}$ indicates the maximum duration.
After processing the gaze information, the two-dimensional coordinates of the fixation points of participants in the trial are obtained. The coordinates are the relative positions of the fixations to the centroid of the shape.

We compare the result of the grasping and viewing tasks through repeated measures analysis of variance (ANOVAs), taking the objects as within-subjects variable and the tasks as between-subjects variable. For all statistical tests, we use 0.05 as the level of significance.

\begin{table*}[thbp]
\centering
\caption{Significant results for two kinds of grasping tasks and viewing tasks repeated measures ANOVAs on the ADF2C, the ADF2I, the ADF2T, and the VAR}
\begin{tabular}{lllll}
\toprule
\textbf{}                                & \textbf{ADF2C} & \textbf{ADF2I} & \textbf{ADF2T} & \textbf{VAR} \\ \hline
\textbf{Vertical grasping and viewing}   &    \textit{F}=0.046, \textit{p} = 0.829           &   \textit{F}=142.699, \textit{p} $<$0.001            &  \textit{F}=5.699, \textit{p}=0.0176                    &  \textit{F}=284.699, \textit{p} $<$0.001           \\ \midrule
\textbf{Horizontal grasping and viewing} &    \textit{F}=3.781, \textit{p} = 0.053           &   \textit{F}=192.511, \textit{p} $<$0.001             &  \textit{F}=164.964, \textit{p} $<$0.001               &  \textit{F}=171.128, \textit{p} $<$0.001            \\ \midrule
\textbf{Grasp and viewing}               &    \textit{F}=3.209, \textit{p} = 0.074           &   \textit{F}=270.596, \textit{p} $<$0.001            &  \textit{F}=80.568, \textit{p}   $<$0.001               &  \textit{F}=437.500, \textit{p} $<$0.001            \\ \midrule
\end{tabular}
\label{ANOVAS}
\end{table*}

\subsection{Number of fixation}
Table~\ref{fixation} shows the average fixation times in one trial. On average, there are 8.18 (\textit{STD}=0.99) fixation points for each grasping task and 8.62 (\textit{STD}=0.76) fixations for each viewing task. The results of repeated measures ANOVA show that the task type affects the number of fixations (\textit{F}=270.60, \textit{p}$<$0.001). But in one trial, the number of fixations will only be an integer, which makes it difficult to distinguish between grasping and viewing by the number of fixations.


\begin{table*}[thbp]
\centering
\caption{Fixations in different trials}
\begin{tabular}{cccccccc}
\toprule
\textbf{Shape}    & \textbf{square} & \textbf{cross} & \textbf{cross\_down} & \textbf{cross\_up} & \textbf{cross\_right} & \textbf{cross\_left}                & \textbf{mean}                  \\ \midrule
\textbf{Viewing}  & 8.71$\pm$0.83   & 8.35$\pm$0.70  & 8.60$\pm$0.89        & 8.57$\pm$0.70      & 8.08$\pm$0.87                     & 8.62$\pm$0.66           &8.62$\pm$0.80           \\ \midrule
\textbf{Grasping} & 8.28$\pm$1.09   & 8.11$\pm$0.69  & 8.64$\pm$1.11        & 8.23$\pm$0.70      & 8.08$\pm$0.36                     & 8.08$\pm$0.36           &8.18$\pm$0.86          \\ \midrule
\end{tabular}
\label{fixation}
\end{table*}

\subsection{Fixations locations}
\begin{figure}[htp]
  \centering
  \includegraphics[width=0.5\textwidth]{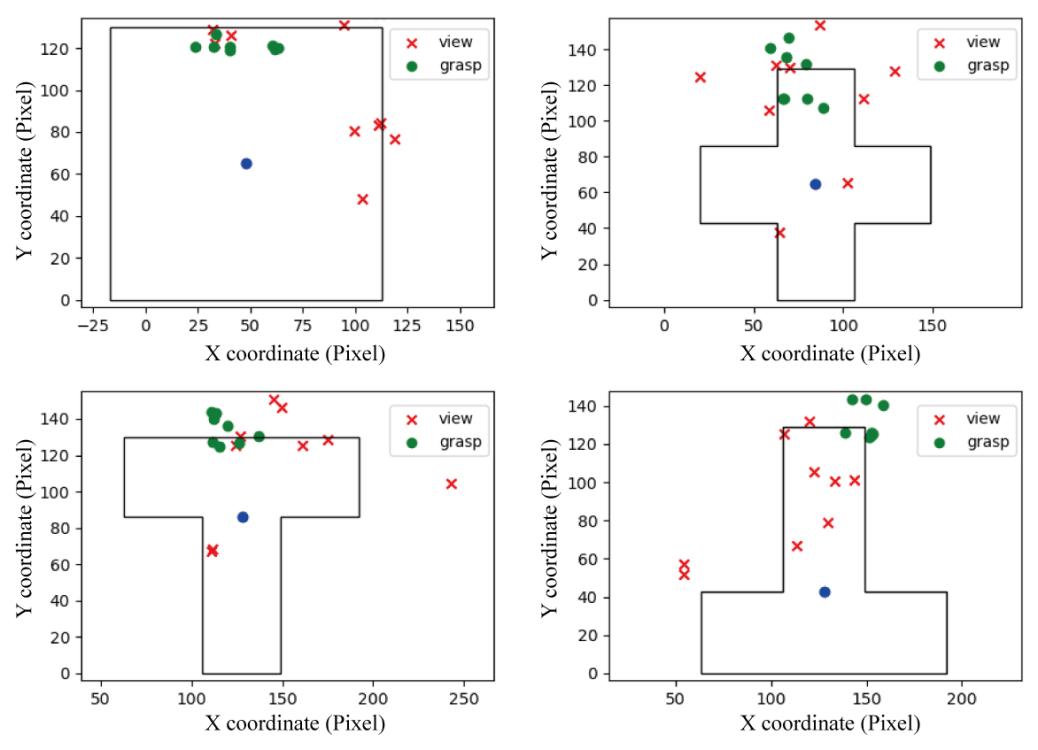}
  \caption{The fixation position of participants on four shapes in certain trials. The centroid of the shapes is represented by the blue point.}
  \label{mean_location}
\end{figure}

Fig.~\ref{mean_location} illustrates the distribution of the fixation location by showing the fixations in certain trials in the grasping and viewing tasks respectively. The centroid of the shapes is represented by the blue point. We can find that fixations have different characteristics under different tasks. To effectively distinguish between grasping and viewing, we propose some features and quantitatively analyze their relationship in different tasks.

In one trial, the user's fixations are expressed as:
\begin{equation}
F({t_i}) = ({f_x}({t_i}),{f_y}({t_i}))
\label{fix_coordinate}
\end{equation}
$i$ represents the number of fixation, and $t_i$ represents the time corresponding to fixation $i$. The two-dimensional centroid coordinates of the object are expressed as follows:
\begin{equation}
C(t) = ({c_x}(t),{c_y}(t))
\label{cog_coordinate}
\end{equation}
where $t$ represents the time of coordinates. The coordinates of the user's thumb and index finger grasp points in the egocentric video are expressed as:
\begin{equation}
\begin{aligned}
{G_1}(t) = (g{1_x}(t),g{1_y}(t))\\
{G_2}(t) = (g{2_x}(t),g{2_y}(t))
\label{grasp_coordinate}
\end{aligned}
\end{equation}

\subsubsection{The average distance from the fixations to the centroid}
The first feature we propose  is the average distance from the user's fixation points to the centroid of the shape (ADF2C). This feature is used to describe the relationship between the participants' fixations and the object:
\begin{equation}
\begin{aligned}
ADF2C = \frac{1}{n}\sum\limits_{i = 1}^n {\sqrt {{{({f_x}({t_i}) - {c_x}({t_i}))}^2} + ({f_y}({t_i}) - {c_y}{{({t_i})}^2}} }
\label{ADF2C_equ}
\end{aligned}
\end{equation}
If the detected centroid does not belong to the user's target object, this feature will appear abnormal (too large), which means that the user's target object can be determined by this feature.

\begin{figure}[thp]
  \centering
  \includegraphics[width=0.5\textwidth]{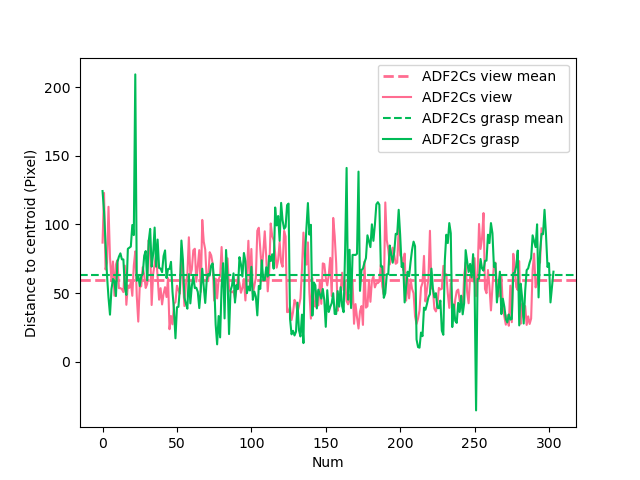}
  \caption{The ADF2Cs in all trials. The solid line represents the result of each trial, and the dotted line represents the average of all results. }
  \label{dis_cog}
\end{figure}

Fig.~\ref{dis_cog} shows the distance between the fixations and the centroid of the shape in grasping and viewing, including the mean value of the distances. Obviously, the curves of ADF2Cs intertwine in the grasping and viewing tasks, and their average values are so close that there is no significant difference.
A repeated measures ANOVA on the ADF2C with the tasks as variable indicates that the ADF2C does not differ between grasping and viewing (\textit{F}=3.20, \textit{p} = 0.07).

\subsubsection{The average distance from the fixations to the grasping point}
\begin{figure}[thp]
  \centering
  \includegraphics[width=0.5\textwidth]{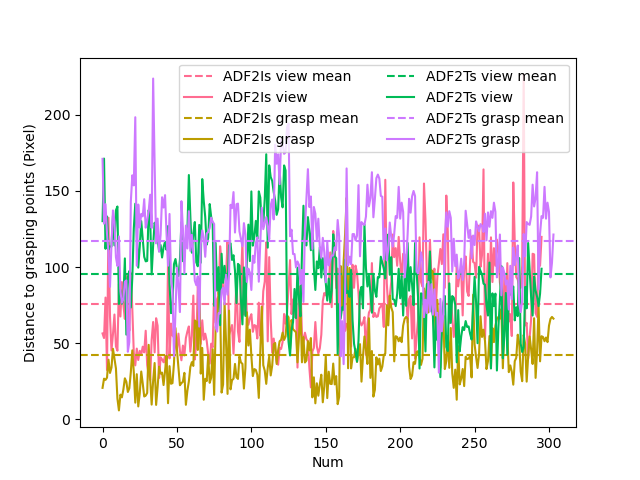}
  \caption{The average distance from the fixation to the grasping point. The solid line represents the result of each trial, and the dotted line represents the average of all results.}
  \label{dis_grasp}
\end{figure}

We find that there are also different relationships between the fixations and the grasp position in different tasks. The grasping positions of the participants are obtained from the pre-experiment. We consider using the average distance between the fixations and the grasping positions in each trial to reflect these relationships.
These two features are the average distance from the fixations to the index finger (ADF2I) and the average distance from the fixations to the thumb (ADF2T):

\begin{equation}
\begin{aligned}
ADF2T = \frac{1}{n}\sum\limits_{i = 1}^n {\sqrt {{{({f_x}({t_i}) - g{1_x}({t_i}))}^2} + ({f_y}({t_i}) - g{1_y}{{({t_i})}^2}} } \\
ADF2I = \frac{1}{n}\sum\limits_{i = 1}^n {\sqrt {{{({f_x}({t_i}) - g{2_x}({t_i}))}^2} + ({f_y}({t_i}) - g{2_y}{{({t_i})}^2}} }
\label{ADF2TI_equ}
\end{aligned}
\end{equation}
Fig.~\ref{dis_grasp} shows the relationship between ADF2I and ADF2T in all trials, respectively. We can find that ADF2I and ADF2T are obviously different in grasping and viewing tasks.
Furthermore, the mean value of ADF2I is significantly lower than that of ADF2T in grasping tasks.
The reason is that in the grasping tasks, the participants' eyes focus on the index finger more than the thumb, which may help stabilize the grasps. A repeated measures ANOVA on the ADF2I shows a significant effect of the tasks (\textit{F}=270.60, \textit{p}$<$0.001). Similar to ADF2I, ADF2T is also affected by the task(\textit{F}=80.57, \textit{p}$<$0.001). Table~\ref{ANOVAS} reveals that this significant effect is more obvious in horizontal grasping tasks than in vertical grasping tasks (the value of \textit{F} indicates the extent of the effect).

\subsubsection{The relationship between fixation points}
We also find that in one trial, there is a relationship between the fixations. We first propose the variance (VAR) of the distances from all the fixation points to the center as a quantitative description of concentration:
\begin{equation}
\begin{aligned}
&O = ({o_x},{o_y}) = (\frac{1}{n}\sum\limits_{i = 1}^n {{f_x}({t_i})} ,\frac{1}{n}\sum\limits_{i = 1}^n {{f_y}({t_i})} )\\
&{d_i} = \sqrt {{{({f_x}({t_i}) - {o_x})}^2} + {{({f_y}({t_i}) - {o_y})}^2}}\\
&M = \frac{1}{n}\sum\limits_{i = 1}^n {{d_i}}\\
&VAR = \frac{{\sum\limits_{i = 1}^n {{{({d_i} - M)}^2}} }}{n}
\end{aligned}
\end{equation}

\begin{figure}[thp]
  \centering
  \includegraphics[width=0.5\textwidth]{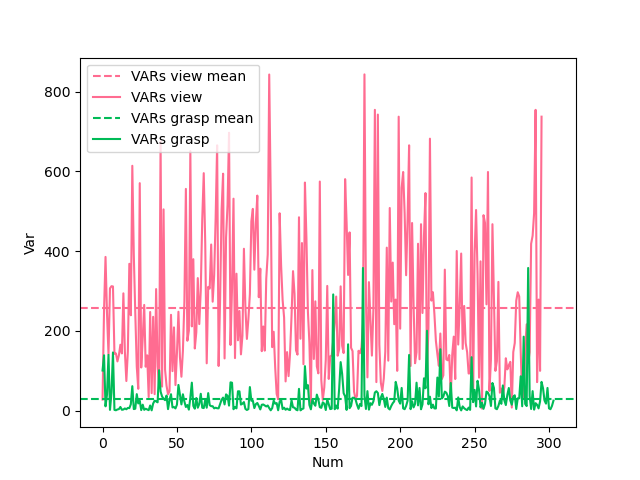}
  \caption{The VARs in all trials. The solid line represents the result of each trial, and the dotted line represents the average of all results.}
  \label{var}
\end{figure}

It can be seen from Fig.~\ref{var} that the VARs in the grasping tasks are significantly smaller than those in the view tasks. In the grasping tasks, the mean value of the VARs is 29.85. \textit{(STD}=44.08), which is much smaller than the viewing tasks (\textit{MEAN}=256.67, \textit{STD}=181.05).
This can be explained by the fact that in the grasping tasks, the participant's attention and sight need to be more concentrated to complete the grasp, and the participant's sight changes randomly across the target during the viewing task.
A repeated measures ANOVA on the VARs show a significant effect of the tasks (\textit{F}=437.50, \textit{p}$<$0.001). This effect is similar in horizontal grasping and vertical grasping.

Table~\ref{ANOVAS} shows all significant results of the repeated measures ANOVA on the ADF2C, the ADF2I, the ADF2T, and the VAR.

\section{Grasping intention recognition}
From the above data analysis, we can know that in the viewing and grasping tasks, participants' fixations have different features, which show the potential of accurately estimating the participants' grasping intention. Thanks to the use of human natural eye behavior, people's cognitive load is reduced and can be quickly accepted by users. This method is especially suitable for patients with upper limb injuries. They can use gaze data to control and express their grasping intention, instead of relying on residual EMG or movement.

To evaluate the effect of different features on grasping intention recognition, we selected five feature combinations.

1) Combination1:ADF2T+VAR

2) Combination2:ADF2I+VAR

3) Combination3:ADF2C+ADF2T+VAR

4) Combination4:ADF2C+ADF2I+VAR

5) Combination5:ADF2C+ADF2I+ADF2T+VAR

Several different machine learning methods are used, including support vector machine (SVM), K-Nearest Neighbor (KNN), stochastic gradient descent (SGD), and decision tree (DT).
The total samples are randomly partitioned into 5 equal-sized subsets. Among the 5 subsets, four subsets are used as the training set while the remaining one is used as the testing set (Test1). For each methods,we use the 5-fold cross-validation to evaluate the accuracy of recognition. The other test set (Test2) is collected from three other shapes that does not exist in the training set, each of which contains 20 samples.

\begin{figure*}[htbp]
\centering
\subfigure[]{

\begin{minipage}[bp]{0.5\linewidth}
\centering
\includegraphics[width=1\linewidth]{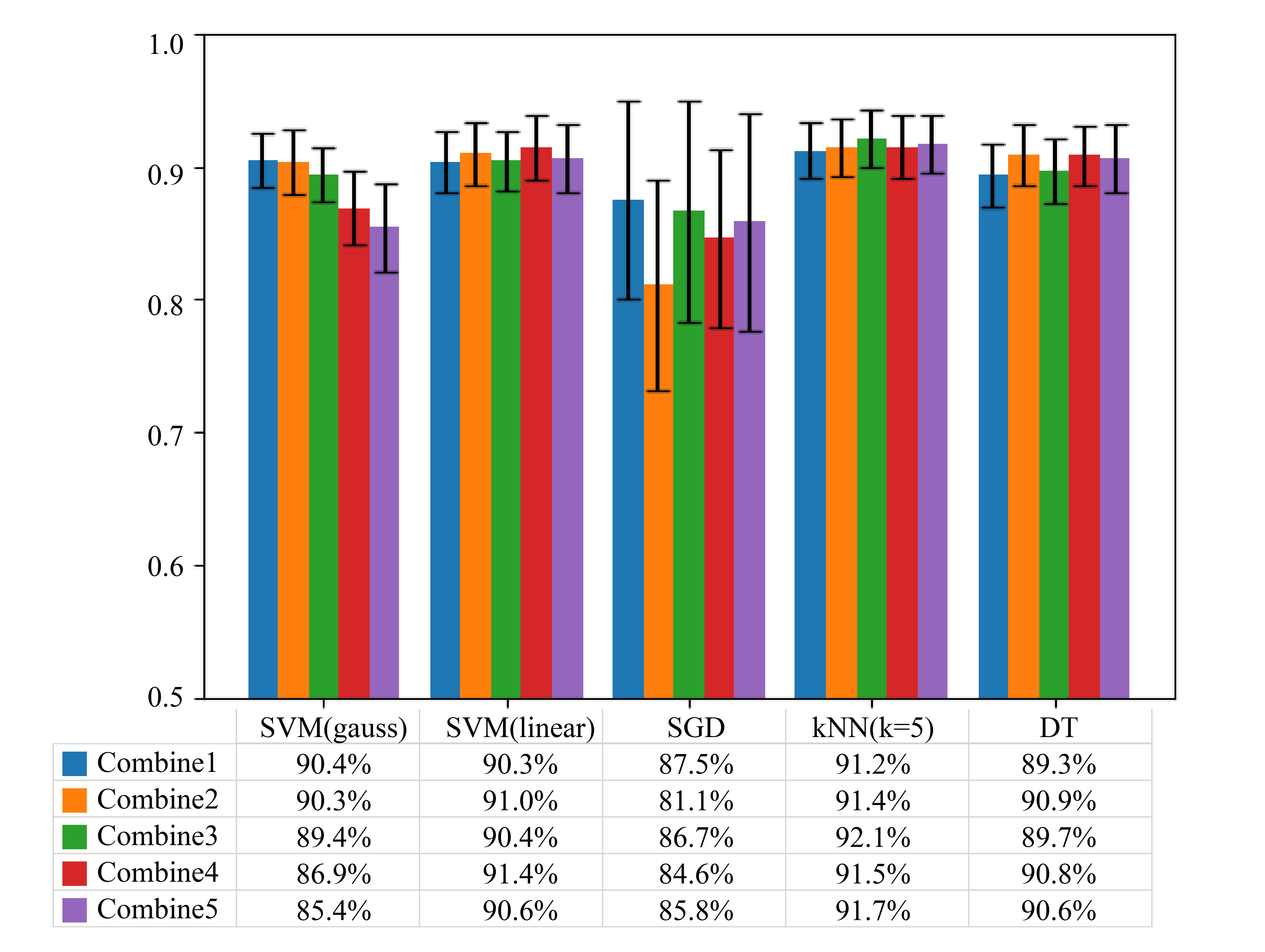}
\end{minipage}
}%
\subfigure[]{
\begin{minipage}[bp]{0.5\linewidth}
\centering
\includegraphics[width=1\linewidth]{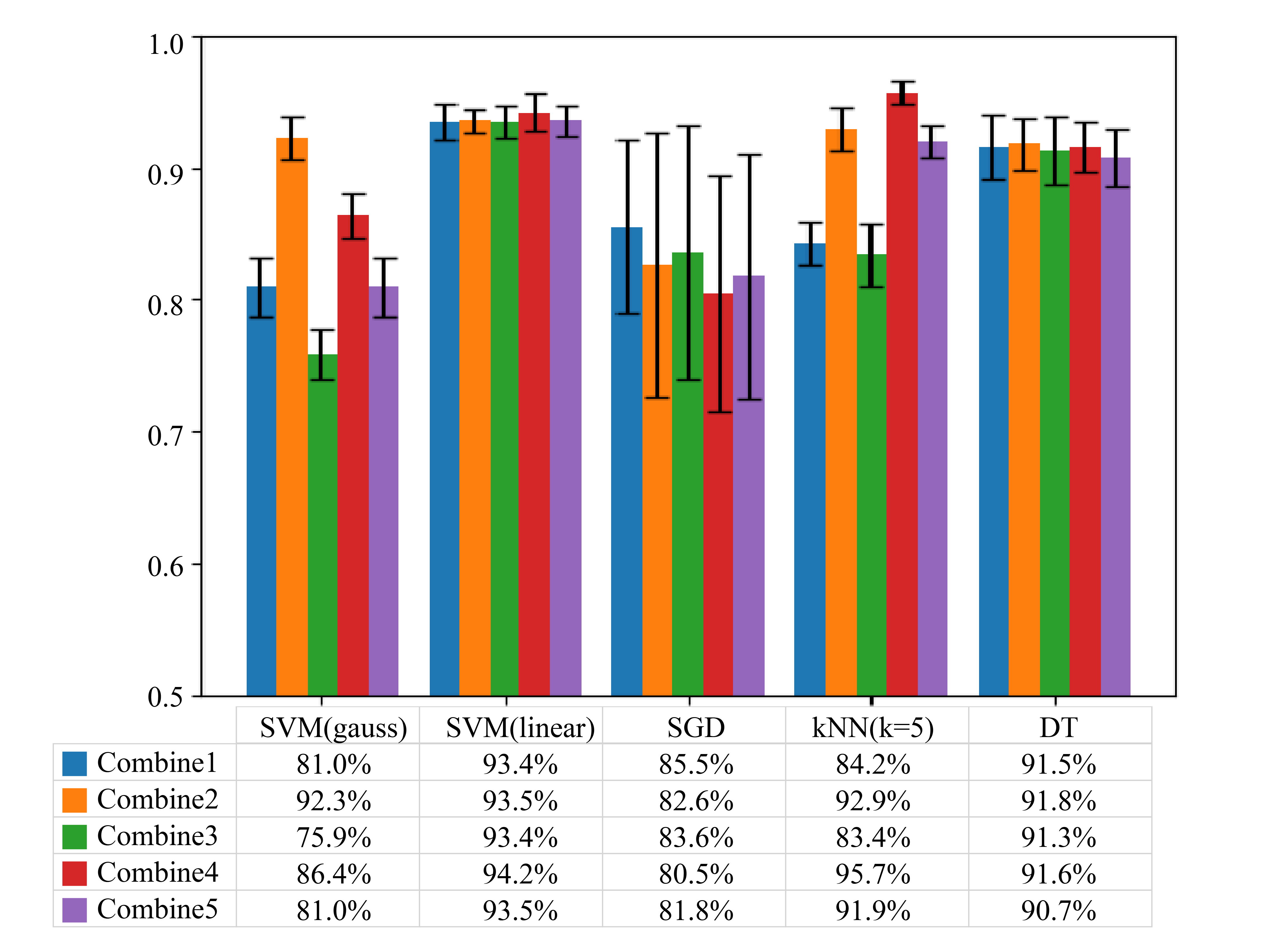}
\end{minipage}
}%

\centering
\caption{Classification accuracy of each classifier and combinations for grasping intention recognition. We perform 100 repeated experiments and used the average accuracy and standard deviation. (a) The accuracy of Test1. (b) The accuracy of Test2}
\label{acc}
\end{figure*}

From Fig.~\ref{acc} we can learn that
for different feature combinations, Combination3 has the highest average accuracy on Test1, which is 89.7\%$\pm$1.9\%. The average recognition accuracy of the other four feature combinations on Test1 also exceeds 88.8\%.
On Test2, Combination2 has the highest average recognition accuracy of 90.6\%$\pm$4.5\%. The average recognition accuracy of other feature combinations is also acceptable, the lowest of which is higher than 87\%.

For different classifiers, the KNN has the highest average accuracy on Test1, which is 91.6\%$\pm$0.3\%. The SGD has the lowest grasping intention recognition accuracy on Test1, with an average of 85.1\%$\pm$2.5\%. But this is also an acceptable result, which means that the influence of classifiers is small.

On Test2, the SVM (linear) achieves the best accuracy of 93.6\%$\pm$0.3\%. The accuracy of other classifiers also exceeds 82\%. The reason why the SVM classifier has higher accuracy on Test2 than Test1 may be that Test1 has a longer experimental process. This makes the participants become fatigued and lead to abnormal fixation mode.

From the analysis, we can know that the features we discovered can effectively identify the user's grasping intention and can achieve excellent results with different classifiers. Furthermore, we have successfully applied the proposed feature combinations to the newly appeared objects in the test set to identify the user's grasping intention. These results show that the features we extracted from the gaze data are robust and can reflect human intention, which can be effectively applied to the recognition of users' grasping intention.

\section{Grasping experiment in real environment}
\begin{figure}[htp]
  \centering
  \includegraphics[width=0.5\textwidth]{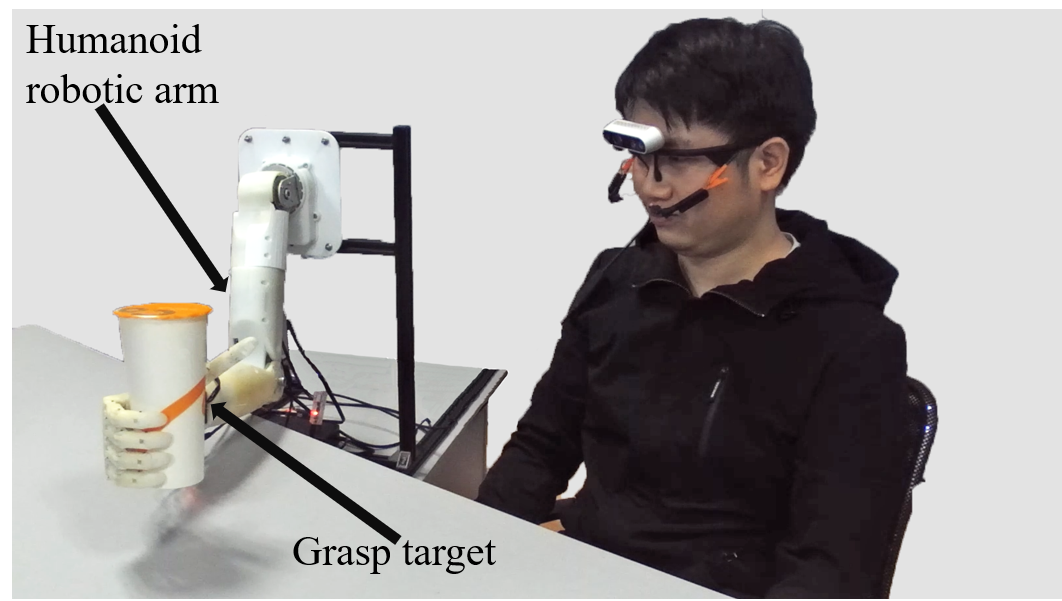}
  \caption{Application experiment in the real environment. The system recognizes the user's grasping intention and controls the humanoid robotic arm to complete the grasp.}
  \label{realexp}
\end{figure}

We have carried out the application experiment in the actual daily scene with the humanoid arm. The experimental scene is shown in the Fig.~\ref{realexp}. We chose a cup that is commonly used in daily life as the intended target. We used the Mask R-CNN~\cite{he2017mask} object detection algorithm to extract the target information we need from the real and complex environment. Through the selected algorithm, we can detect the mask of the cup from the user's egocentric scene and use it to calculate the two-dimensional centroid of the object.
We set up a sliding window with a duration of three seconds to sample the data. Then, the features are extracted by combining the centroid of the object, the grasping points with the user's fixation information. At the same time, we use the Combination4 and the KNN as features and the classifier for grasping intention recognition. The experimental results show that, combined with the current advanced image processing algorithms, our proposed method can effectively identify the user's grasping intention and successfully implement the grasp. The complete experimental process is shown in the video in the material.

Four subjects participated in our experiment. To complete the objective evaluation, the subjects were asked to complete the Quebec User Evaluation of Satisfaction with assistive Technology (QUEST) questionnaire~\cite{demers2002quebec}. The results are shown in Table~\ref{question}. From the QUEST questionnaire we find that although there is general agreement with effectiveness, easiness to use, and safety, there is displeasure on comfort. This is probably because the scene camera is relatively heavy and the eye-tracker is not a mature wearable  product.

\begin{table}[htbp]
\centering
\caption{QUEST evaluation results for the humanoid robotic arm assist system}
\begin{tabular}{ccccc}
\hline
\textbf{}                       & \textbf{Safety} & \textbf{\begin{tabular}[c]{@{}c@{}}Easy \\ to Use\end{tabular}} & \textbf{Comfort} & \textbf{\begin{tabular}[c]{@{}c@{}}Effec\\ -tive\end{tabular}} \\ \hline
\textbf{Not satified at all}    & 0               & 0                    & 0                & 0                  \\ \hline
\textbf{Not very satisfied}     & 0               & 0                    & 1                & 0                  \\ \hline
\textbf{More or less satisfied} & 1               & 0                    & 1                & 0                  \\ \hline
\textbf{Quite satisfied}        & 3               & 3                    & 2                & 2                  \\ \hline
\textbf{Very satisfied}         & 0               & 1                    & 0                & 2                  \\ \hline
\end{tabular}
\label{question}
\end{table}

\section{Discussion and Conclusion}
Recognition of grasping intentions is the premise for wearable robots to assist users in grasping, but in some cases, the eye movements are the only intention related activities that can be observed. So we study the eye movements related to grasping.

We obtain the fixation messages of participants in grasping and viewing tasks through experiments. These messages are combined with the target centroid and the grasping positions to study the fixation laws under different tasks. Through the experimental results and statistical analysis, we find that fixations are significantly different in the two tasks.
In the grasping task, the participants' fixations are closer to the grasp position of the index finger and concentrate on a small area. While in viewing tasks, the distribution of participants' fixations are arbitrary, which may be distributed in multiple parts of the target.
This phenomenon has inspired us to use these natural gaze fixations to identify the grasping intention of participants.

We first propose four features- the ADF2C, the ADF2I, the ADF2T, and the VAR to describe these differences quantitatively. Utilizing the quantitative features, we can use machine learning methods to recognize the user's grasp intention, and further control the robot to help the user complete the grasp.
It is crucial that only the natural gaze fixations of the human are used in our method, and no additional actions or commands are required. Thus, this method is very suitable for patients with hemiplegia.
However, not all users' fixations follow the rules we have found. Some users' fixations are close the centroid or the grasp point of the thumb rather than the grasp point of the index finger when grasping. When the user knows the pose and shape of the object in advance, he/she may be able to grasp it without staring at it.
Additionally, the work of this paper explores the relationships of the users' fixations in different tasks when the thumb and index finger are visible.
For some grasping tasks, the contact position of the index finger is not visible. In the next work, we will explore the laws of fixations in grasping and viewing tasks when not all the grasping points are visible. Furthermore, we intend to apply the method to patients with hemiplegia so that they can complete daily grasping tasks.

In conclusion, we are exploiting the clues behind eye movements, which are highly correlated to our motor actions. They are key factors in human motor planning and imply our movement intention. Consequently, gaze tracking becomes a rich source of information about user intent that can be utilized. Fixations reflect the focus of human gaze. We have proved that only a few fixations are needed to quickly and accurately recognize the user's grasping intention. This technology bears a tremendous potential as an interface for human-robot interaction. Moreover, eye-movements are retained in the majority of motor disabilities, which enhances the applicability of eye-tracking technology in the field of human-robot interfaces for the paralysed and in prosthetics. Unlike EMG based and EEG based methods that require long-term training, our method only needs natural eye movement and does not need training. The intention recognition performance of our system are subject to the accuracy of gaze estimation. We hope to improve the accuracy of gaze estimation in the future to achieve more accurate recognition results. Our system allows for straightforward control of the robot for grasping. The application in the real environment has proved its effectiveness. Furthermore, the user's fixations can also provide the location information of the target. If we combine the user's intention with the position information of the object, the user can control the humanoid robotic arm to grasp the object with his eyes only. The ability to recognize the user's grasp intention from fixations enables novel ways of achieving the embodiment in robotic solutions that restore or augment human functions.


%
%
\addtolength{\textheight}{-12cm}   

\bibliographystyle{IEEEtran}
\bibliography{mylib}

\end{document}